Fraser Robinson*, Souren Pashangpour*, Matthew Lisondra*, Goldie Nejat

*Autonomous Systems and Biomechatronics Laboratory, Department of Mechanical and Industrial Engineering, University of Toronto, Toronto, Canada*

This research is supported by AGE-WELL Inc., an NSERC CREATE HeRo fellowship and the Canada Research Chairs Program.

F. Robinson, S. Pashangpour, M. Lisondra, and G. Nejat are with the Autonomous Systems and Biomechatronics Laboratory of the Department of Mechanical and Industrial Engineering, University of Toronto, Toronto, ON M5S 3G8 Canada.

(e-mail: fraser.robinson@mail.utoronto.ca, souren.pashangpour@mail.utoronto.ca, matthew.lisondra@mail.utoronto.ca, goldie.nejat@utoronto.ca).

* Authors contributed equally to this research.

# PovNet+: A Deep Learning Architecture for Socially Assistive Robots to Learn and Assist with Multiple Activities of Daily Living

F. Robinson*, S. Pashangpour*, M. Lisondra*, and G. Nejat

A significant barrier to the long-term deployment of autonomous socially assistive robots is their inability to both perceive and assist with multiple activities of daily living (ADLs). In this paper, we present the first multimodal deep learning architecture, POVNet+, for multi-activity recognition for socially assistive robots to proactively initiate assistive behaviors. Our novel architecture introduces the use of both ADL and motion embedding spaces to uniquely distinguish between a known ADL being performed, a new unseen ADL, or a known ADL being performed atypically in order to assist people in real scenarios. Furthermore, we apply a novel user state estimation method to the motion embedding space to recognize new ADLs while monitoring user performance. This ADL perception information is used to proactively initiate robot assistive interactions. Comparison experiments with state-of-the-art human activity recognition methods show our POVNet+ method has higher ADL classification accuracy. Human-robot interaction experiments in a cluttered living environment with multiple users and the socially assistive robot Leia using POVNet+ demonstrate the ability of our multi-modal ADL architecture in successfully identifying different seen and unseen ADLs, and ADLs being performed atypically, while initiating appropriate assistive human-robot interactions.

Keywords: Socially assistive robots; human activity recognition; activities of daily living; multimodal deep learning; sensor fusion; RGB-D perception; user state estimation; proactive human-robot interaction; aging-in-place

## Introduction

Numerous conditions including neurological, musculoskeletal, sensory and/or circulatory can reduce a person's capacity to independently complete activities of daily living (ADLs) [1], [2]. Furthermore, aging is a significant factor in ADL impairment [3]. As dependence on ADL support increases, individuals may require home-care [4] or relocation to long-term care (LTC) homes [5], both of which can cause feelings of dependence and isolation, decreasing a person's quality of life (QoL) [5]. A proposed care method for ADL assistance while maintaining QoL and independence is reablement [6].

Reablement is a person-centered approach for gaining or regaining skills required to complete ADLs [6]. It consists of goal-directed rehabilitation interventions in which caregivers coach through encouragement and motivation to expand individual's capabilities [7]. Deployment of reablement programs have shown significant improvements in health and ADL ability in addition to decreased care costs for a variety of individuals including older adults [8], [9] and stroke patients [7]. However, a global shortage of caregivers has limited the implementation of these programs [10], and assistive technologies have not yet been incorporated.

Socially assistive robots (SARs) have been used to assist individuals with ADLs including dressing [11], exercise [12], or meal eating and cognitive games [13]. For SARs to engage in assistive human-robot interactions (HRI) for reablement, they must initiate interactions. In general, initiation of HRI with SARs has been achieved by either the user, a caregiver, or the robot itself [14]. We define the first two types as "reactive" HRI as a human must initiate the interaction, whereas robot-initiated HRI is "proactive" and requires the SAR to autonomously determine when to initiate assistance. SARs capable of perceiving and acting upon human behaviors to initiate HRI can promote long-term use [15] and improve ease of use [16]. Proactive HRI for multi-activity assistance requires SARs to differentiate between numerous ADLs. Thus, both offline learning to detect common ADLs, and online learning to recognize and learn new ADLs is needed. In particular, online learning can be achieved on a group of low-dimensional activity representations known as an embedding space [17].

Existing ADL recognition and monitoring architectures used in SARs classify user motion based on an ADL class which has been already defined during the offline learning phase [18]. However, they are not able to isolate non-ADL movements (e.g., walking between rooms) resulting in false positive classification of these movements as an ADL class. False positives can result in SARs providing incorrect assistive behaviors which may confuse users and decrease their perception of SAR capabilities [19]. Furthermore, to-date, SARs for multi-activity assistance are not capable of recognizing new activity classes online [20]-[23].

In this paper, we propose the development of a framework for autonomously identifying and classifying ADLs to provide *multi-activity proactive* HRI by SARs. This research builds upon our previous work in [24] where we developed the first deep learning ADL classifier and embedding space framework using multimodal spatial-mid fusion. The method, however, requires users to implement specific motions related to an ADL class. Therefore, if a user performs non-ADL movements, these movements will result in forced classification into a *seen* ADL rather than being excluded from the embedding space, resulting in a false positive. We extend our previous research work to robustly classify multiple different classes to account for non-ADLs and varying user preferences/limitations. Our novel contributions include: 1) the development of a new user motion embedding space which provides a low-dimensional representation of user joint motion to isolate ADL motion from non-ADL motion, 2) the development of a novel user state estimation method that uniquely applies a similarity function on the ADL embedding space to identify *unseen* ADLs and recognize ADLs performed atypically, and 3) the first implementation and validation of a multimodal deep learning (DL) architecture for autonomous multi-activity assistance by a SAR.

**Related Work**

This section provides a review of the related work on SARs, including: 1) learning methods for classifying ADLs, and 2) user state tracking for ADL assistance.

*Classification of ADLs for SARs*

Recognizing and classifying multiple ADLs is essential for autonomous SAR assistance [25]. Early approaches primarily relied on unimodal ML methods using 3D joint data and handcrafted features, such as support vector machines-, k-nearest neighbour-, and hidden Markov model-based models, which required predefined feature engineering and were limited in generalization across users and environments [26], [27]. To overcome these constraints, recent works have increasingly adopted DL architectures capable of learning motion and contextual features directly from data [28], [29]. For example, DL-based methods have used RGB video and pose inputs to recognize user intent and task stages in assistive scenarios including drink preparation, personal hygiene, and upper-body activities, demonstrating improved online and real-time capabilities on SAR platforms. More recent advances emphasize multimodal fusion, combining RGB, skeleton, and contextual object information to capture complementary semantics and improve robustness to user variability and environmental clutter. For example, fusion architectures that learn modality-specific attention (e.g., ESE-FN [30]) and recent SAR-oriented multimodal datasets that better reflect in-home deployment constraints (e.g., privacy-preserving PriMA-Care [31]).

*User State Estimation for ADL Assistance*

User state estimation in ADL contexts is typically modeled as task progress across discrete states describing the steps required to complete an activity [32]. Within SAR systems, these states have been inferred using three main classes of methods: rule-based, classical machine learning (ML), and DL. Rule-based approaches define user states through expert-specified thresholds applied to sensed behavioral or physiological signals. These have been deployed for exercise coaching [33], cardiac rehabilitation [34], and meal-time assistance [35], where states such as correct/incorrect execution, engagement, or activity completion are inferred from posture, heart rate variability, electromyography, utensil interactions, or facial orientation [36].

To reduce reliance on handcrafted thresholds, ML-based methods have employed data-driven classification of user motion features. For instance, K-NN classifiers have been used to classify exercise completion levels using 3D joint positions [37]. More recent work includes DL-based estimation, where CNNs and vision-based detection models infer user state directly from sensor data [38]. These approaches have been applied to dressing assistance via wearable strain sensing [39], eating behavior monitoring [40], and handwashing step recognition [41] using RGB imagery, demonstrating improved robustness and generalization in real-time robotic assistance [42].

Research has also focused on adaptive user state estimation, where task progress is inferred without fixed state definitions or dense labeling [43]. For example, the PACE framework [44] estimates user task progress via dynamic time warping and then trains a reinforcement learning agent to synchronize assistive actions with the user's behavior using only sparse demonstrations.

*Summary of Limitations*

The existing ADL classification learning methods developed for SARs classify discrete human activity classes based on user/object motion and/or scene background features. DL methods improve upon ML methods by learning optimal features for increased classification accuracy. However, there has been limited research on utilizing multimodal data specifically for multi-activity recognition for SARs to proactively initiate assistive behaviors, which can increase classification accuracy compared to unimodal approaches by extracting rich complimentary semantic features [45]. Furthermore, existing ADL classification methods do not determine when users are performing non-ADL movements, as they require all inputs to be from known activity classes and are not able to isolate inputs which are not defined prior to classification. User state estimation methods for ADL assistance categorize user states as discrete steps in a task specific sequence. However, these methods are not able to identify *unseen* activities or known activities that are being performed atypically by users. Namely, they require labeling of data and offline training to classify new ADLs or atypical ADL performance.

This work extends our multimodal deep learning ADL classifier [24] by incorporating a novel motion embedding space that identifies and isolates ADL movements during implementation. This addresses the limitation of needing to predefine inputs prior to classification. We also propose a user state estimation method which recognizes new *unseen* ADLs or *atypically performed* ADLs by using an ADL similarity metric for observed unlabeled ADLs in real-time, without requiring labeled data for these ADL types.

# A Deep Learning Architecture for Autonomous Multi-ADL Recognition and Monitoring

The objective of our DL architecture is to enable a SAR to autonomously initiate assistive HRI for a variety of known and unknown ADLs, Fig. 1. RGB-D videos of a user performing an ADL in an environment is provided to the *Multimodal Sampling Module* in order to segment the three modes of RGB video used for scene and motion features, 2) RGB image of both the user and environment for object detection, and 3) 3D pose skeleton sequence of the user for extracting spatially independent user motion features. The *ADL Classifier Module* uses these input modes to generate vector $\vec{ADL}$, which contains: 1) the corresponding ADL task class (specific user related ADL-actions) vector $\vec{T}$, and 2) a pair of embedding vectors, ADL embedding vector $\vec{e}_o$ and user motion embedding vector $\vec{M}_u$. The embedding vectors are used by the *User State Estimation Module* to classify the specific ADL type (*seen*, *unseen*, or *atypically performed*), and ADL task class within the ADL type. ADL type and task class are used by the robot to monitor the user and determine the appropriate socially assistive behaviors to provide. Herein, we define our pose, object, and video network DL framework as POVNet+.

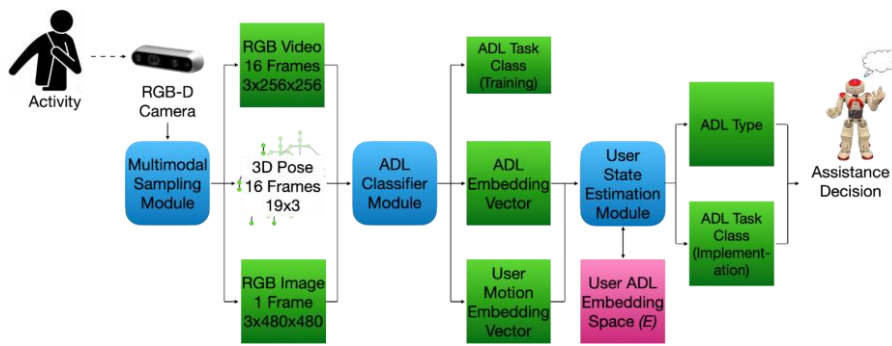

Figure 1. Deep Learning Architecture POVNet+ for ADL Monitoring and Assistance during Proactive HRI.

*Multimodal Sampling Module*

The Multimodal Sampling Module takes as input an RGB-D video stream of a user performing an ADL in a scene and uses a combination of skeleton pose tracking and rolling window sampling of the RGB-D video to provide a multimodal output of RGB video, RGB image, and 3D user pose. Skeleton pose tracking uses the Nuitrack API [46] to obtain the 3D poses of 19 joints on the body representing the motion of the head, neck, collar, waist, shoulders, elbows, wrists, hands, hips, knees, and ankles for identifying ADLs. Rolling window sampling is applied at a frequency of 6 Hz on the 30 FPS video input such that five frames of RGB video and 3D user joint positions are collected per second. A final output of 1) 16 frames of $3 \times 256 \times 256$ video, 2) 16 frames of $3 \times 19$ 3D joint positions, and 3) 1 RGB image of size $3 \times 480 \times 480$ is provided to *ADL Classifier Module*.

*ADL Classifier Module*

The objective of the *ADL Classifier Module* is to learn task-discriminative latent representations from the normalized multimodal inputs generated by the *Multimodal Sampling Module*. This module maps synchronized video, pose, and object features into a shared representation for ADL task classification and downstream user state estimation. Modality-specific backbone networks are used for feature extraction, followed by spatial mid-fusion layers that integrate complementary motion, object, and scene information, Fig. 2. During training, the resulting representation is used to predict the ADL task class, and the learned embedding vectors are passed to the *User State Estimation Module* for identifying seen, unseen, and atypically performed ADLs. We discuss each backbone network and our spatial mid-fusion method in more details below.

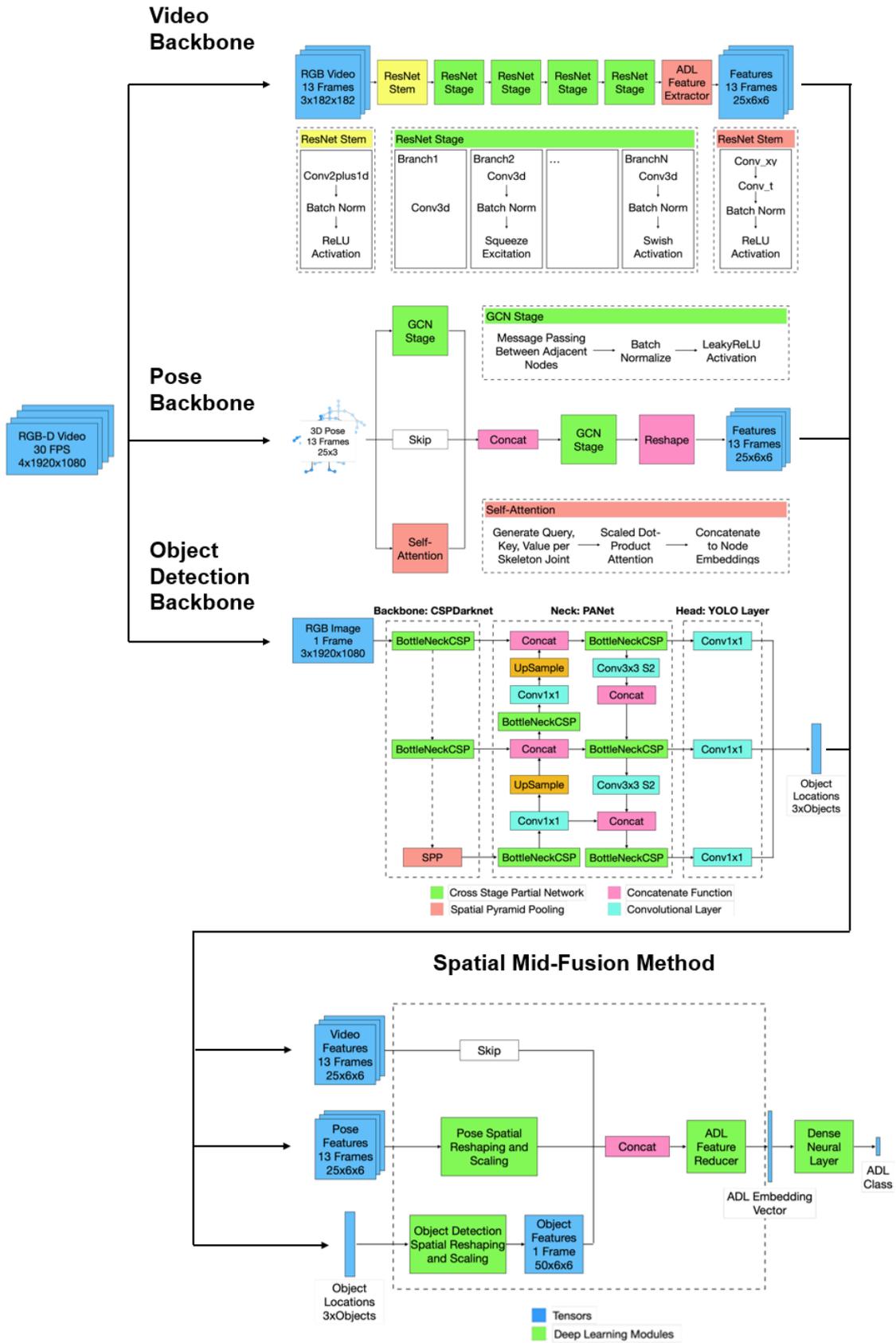

Figure 2. The *ADL Classifier Module* including the Video, Pose and Object Detection Backbones and the Spatial Mid-Fusion Method.

*Video Backbone*

The video backbone extracts scene background and user motion features using a X3D-m video recognition architecture [47] followed by an ADL feature extraction 3D convolutional layer, Fig. 2. X3D-m is based on an expanding ResNet architecture which optimizes dimensions of temporal duration, frame rate, spatial resolution, width, bottleneck width, and depth for accuracy complexity trade-off. The final output of the video backbone is 16 timesteps × 19 channels × (6 × 6) spatially dependent feature grids.

*Pose Backbone*

The pose backbone extracts user motion features that are scene and scale invariant using two graphical convolutional network (GCN) stages [48], Fig. 2. The input 3D skeleton pose sequence is represented as a graph data structure where the nodes are the 3D positions (*x,y,z*) of each skeleton joint and the edges are the skeleton structure of Nuitrack. The graph data structure can be represented as $G = (V, E, X)$, where $G$ is the graph representation of the 3D skeleton pose, while $V$ is the set of nodes which represent the joints. $E$ represent the edges, defining the skeletons connectivity, and $X$ represents the 3D positions of each joint.

The first layer of the backbone uses a: 1) GCN stage for localized motion features, 2) a self-attention layer for global motion features, and 3) skip connection for graph constrained motion features. These parallel branches are concatenated and used in the next GCN stage for joint variant motion features.

A motion embedding vector extraction technique is also utilized by the pose backbone to represent the distance that each joint moves throughout the observed user motion while performing the ADL. The Euclidean distance traveled $D$ for each joint $J$ across discrete timesteps $t = 0\ to\ t = T$ is defined as:

$$D = \|J_{0 \to T}\| = \|J_0 - J_1\| + \cdots + \|J_{T-1} - J_T\|. \tag{1}$$

The output of the pose backbone includes: 1) user motion features for spatial mid-fusion, in the shape *16 timesteps × 19 channels (one per node) × (6 × 6) spatially independent feature grids*, and 2) A user motion embedding vector $\vec{M}_u$ of size 19, where each element $M_{u,j}$ represents the total Euclidian displacement of a specific joint $j$ over the observation window $W$. This is computed as: $M_{u,j} = \sum_{=0}^{W} \|J_{j,t+1} - J_{j,t}\|$, where $J_{j,t}$ represents the 3D position of joint $j$ at time $t$. Higher values indicate greater movement for specific joints, while lower values correspond to smaller localized motions. This representation enables the system to differentiate between fine-motor ADLs (e.g., taking medicine) that involve limited, localized hand motions and gross-motor ADLs (e.g., standing up from a chair) that require large, coordinated movements of the arms, legs, and torso.

*Object Detection Backbone*

The object detection backbone identifies and localizes ADL-specific objects in the scene (e.g., cup, brush, chair). We use the YOLOv13 [49] object detection model to identify the 2D location of the centroid of the bounding box for each identified object in the RGB image of the scene. Thirty-eight household objects that are used in ADLs were selected from the COCO dataset [50] and used to pretrain the network. Object locations are provided as an output in the form of *1 timestep × 38 possible objects × number of objects detected × 2 object locations (x, y)*.

*Spatial Mid-Fusion*

The spatial mid-fusion method is used to combine and synthesize multimodal features from each modality into a unified spatial understanding of a user's environment. Our spatial mid-fusion method developed in [24] uses the spatial reference from the video backbone, which defines absolute 2D spatial coordinates relative to the video boundaries, to unify spatial understanding for the other modalities. Spatial understanding is applied to each pose feature grid *F*, size *6 × 6*, for all *19 user joints* and *16 timesteps* by matrix multiplication with a *6 × 6* matrix describing the absolute spatial position of the joint at that timestep. For the object backbone output, spatial understanding is applied to each of the *38* possible objects to define a *6 × 6* matrix describing the absolute spatial location of each object instance for an output of *1 × 38 × 6 × 6*. Outputs from each backbone are concatenated before the fusion layer for a size of *17 × 38 × 6 × 6*.

The fusion layer is a spatial 2D CNN layer followed by a temporal 1D convolutional layer to extract ADL features from the fused multimodal data. The output of the spatial mid-fusion layers is the ADL embedding vector of size *128* used in the *User State Estimation Module*. Additionally, for training the network, a dense neural layer is then used to classify the ADL based on the ADL embedding vector using spatially dependent linear weights and non-linear activations.

By synthesizing multimodal information, the SAR can interpret complex activity contexts, such as identifying eating when it observes both hand-to-mouth motion and the presence of utensils. This approach reduces the likelihood of false positives and improves ADL classification accuracy by ensuring that detected actions and objects are spatially aligned.

*User State Estimation Module*

The *User State Estimation Module* identifies: 1) the user motion type (e.g., ADL or non-ADL), 2) the ADL type (e.g., *seen*, *unseen*, *atypical*), and 3) the ADL task class (e.g., flossing, putting on hat, etc.). The inputs to the module are the ADL and user motion embedding vectors.

*ADL or Non-ADL Detection*

An ADL motion isolation method is used to determine whether the observed motion is an ADL or non-ADL motion. The user motion embedding vector $\vec{M}_u$, generated by the Pose Backbone, contains the cumulative motion of the 19 joints. The average motion of all joints is:

$$\overline{\vec{M}_u} = \frac{\sum_{i=0}^{19} \vec{m}_i}{\text{len}(\vec{M}_u)}, \qquad (2)$$

where $\vec{m}_i$ is the motion of a given joint *i*. Averaging filters out noise from individual joint variations or minor fluctuations in movement. $\overline{\vec{M}_u}$ is determined to be an ADL (*seen*, *unseen*, or *atypically performed*):

$$\text{if } |\vec{M}_{min}| < |\overline{\vec{M}_u}| < |\vec{M}_{max}|, \text{ else non-ADL}, \qquad (3)$$

where $|\vec{M}_{min}|$ and $|\vec{M}_{max}|$ are scalar values representing the minimum and maximum average global body motion observed during ADL execution. These values are determined from training data by computing the distribution of the averaged joint-motion measure $|\overline{\vec{M}_u}|$ across all recognized ADL instances. $|\vec{M}_{min}|$ represents the minimum motion intensity required for an observed behavior to be considered an ADL, thereby filtering out non-ADL movements such as fidgeting, posture adjustments, or brief transitional motions.

$|\vec{M}_{max}|$ defines the upper bound of expected motion intensity; exceeding $|\vec{M}_{max}|$ can indicate atypical task execution or non-task-related movements.

The output of the ADL motion isolation method is a Boolean user motion type $M_{ADL}$, 1 for ADL and 0 for non-ADL motion. If ADL motion is detected, the system proceeds to ADL classification using the full joint-level motion representation. Specifically, the *ADL Classifier Module* generates a compact ADL embedding vector $\vec{e}_o = [M_u, o_{ADL}]$ where $M_u$ preserves detailed joint displacement information over time, computed as:

$$M_{u,j} = \sum_{=0}^{T} ||J_{j,t+1} - J_{j,t}||, \tag{4}$$

and $o_{ADL}$ encodes the object interactions and spatial features extracted via spatial mid-fusion. This structured embedding enables direct comparison between ADL instances using Euclidean distance in the embedding space $E$.

*Seen, Unseen or Atypical ADL*

The ADL type of an observed activity is classified using a scalar similarity score $S$, computed in the learned ADL embedding space $E$. The embedding space $E$ contains ADL embedding vectors $\vec{e}_o$ and their associated ADL task labels. $E$ is initialized using ADL embeddings from the training dataset. During online implementation, $E$ is incrementally updated for each user with newly observed ADL embeddings when $M_{ADL} = 1$.

For each ADL class $i$, a centroid $C_i$ is computed as the mean of all embeddings belonging to that class:

$$C_i = \frac{\sum_{j=0}^{n_i} \vec{e}_j}{n_i}. \tag{5}$$

Centroids are updated when a new embedding of the corresponding class is added to $E$. Given an observed ADL embedding $\vec{e_o}$, the similarity to known ADL classes is measured using:

$$S = \frac{D_{min}}{D_k} \times \frac{1}{var_k}, \qquad (6)$$

where $D_{min}$ is the minimum Euclidian distance between $\vec{e_o}$ and all centroids $C_N$:

$$D_{min} = \min(\|\vec{e_o} - C_0\|, \dots, \|\vec{e_o} - C_N\|), \qquad (7)$$

and $k$ denotes the class achieving this minimum distance. $D_k$ is the mean intra-class Euclidean distance between embeddings of class $k$ and centroid $C_k$:

$$D_k = \frac{\sum_{j=0}^{n_k} \|\vec{e_{k,j}} - C_k\|}{n_k}. \qquad (8)$$

$var_k$ is the intra-class variance of ADL embeddings belonging to class $k$. Variance normalization is applied to account for ADL task classes with greater intra-class variance. A lower $S$ represents higher similarity between ADLs. $S$ values for each observed activity are stored for a given user $u$ in the set $\{S_u\}$ for detecting *atypically performed ADLs*. Unlike conventional classifiers that output only discrete class labels, our approach uses the learned embedding space to apply a distance-based requirement to determine whether an observed ADL instance is typical or atypical for its predicted class.

**Training of POVNet+**

The POVNet+ was trained to learn ADL scene, motion, and object features in order to construct a user ADL embedding space $E$. For training, ADL-specific samples performed by older adult subjects were selected from the ETRI-Activity-3D dataset (ETRI) [51], enabling the model to learn a multimodal representation that jointly captures user motion, scene context, and object interactions. A summary of the dataset composition, training protocol, optimization settings, and evaluation setup is provided in Table 1.

TABLE 1. TRAINING, FINE-TUNING, AND EVALUATION SETUP FOR POVNET+

| Aspect | Details |
|---|---|
| Dataset | ETRI-Activity-3D (ETRI) |
| Subjects | 100 total (50 older adults, 50 younger adults) |
| ADL Classes Used | 11 ADLs (older-adult subset) |
| Samples Used | 11,892 (training); 627 (out-of-domain fine-tuning) |
| Input Modalities | RGB video, depth maps, 3D human pose |
| Train/Test Protocol | Cross-subject split ($\approx$ 2/3 train, 1/3 test) |
| Validation Split | 70% / 30% (within training set) |
| Optimization | Cross-entropy loss, learning rate $2 \times 10^{-4}$, batch size 128, 20 epochs |
| Fine-Tuned Components | Pose backbone and spatial mid-fusion layers |
| Fine-Tuning Purpose | Adaptation to new users, environments, and different skeleton tracking APIs |
| Models Used | Untuned model for SOTA comparisons; tuned model for HRI experiments |
| Evaluation Metrics | Precision, Recall, *F1 score* |
| Hardware | RTX 3090 GPU, Ryzen 7 7700X CPU, 64 GB RAM |

Following offline training, selective fine-tuning was applied to the pose backbone and spatial mid-fusion layers to improve robustness to domain shifts arising from variations in 3D skeleton tracking accuracy across different APIs, as well as changes in the spatial and semantic relationships between scene, user motion, and object features introduced by new environments. Fine-tuning was performed using a custom out-of-domain dataset collected from new users, together with a small subset of the original training data to mitigate catastrophic forgetting while preserving previously learned ADL representations.

For experimental evaluation, the untuned POVNet+ model was used for state-of-the-art (SOTA) comparison experiments to ensure a general, non-task-specific assessment, whereas the tuned model was deployed in real-time HRI experiments to account for domain shifts encountered during real-world operation.

**Comparison Testing**

We compared our POVNet+ method to SOTA ADL classification techniques under two evaluation settings: 1) cross-subject testing using within-domain data with unseen users, and 2) out-of-domain testing using unseen users in a new environment. Namely, POVNet+ was benchmarked against the following five unimodal and multimodal ADL recognition models: 1) SlowFast [52], 2) mmaction2 [53], 3) ST-GCN [48], 4) MSAF [54], and 5) POVNet [24]. To the best of our knowledge, POVNet+ is the only method in this comparison which integrates RGB video, 3D pose, and object location information within a single end-to-end architecture.

**SlowFast (unimodal)** [52]: SlowFast is an action recognition architecture which uses a two-pathway 3D CNN in which a slow pathway processes sparsely sampled RGB video frames to capture spatial semantics, while a lightweight fast pathway processes at a higher frame rate to capture motion at a fine temporal resolution; the two pathways are fused through lateral connections. For classification, a global average pooling operation is applied to the output of each pathway, the pooled feature vectors are concatenated, and the resulting representation is passed to a fully connected classifier layer to predict the ADL class from a predefined set of known activities (closed-set recognition).

**mmaction2 (unimodal)** [53]: mmaction2 is an action recognition framework that provides standardized implementations of spatio-temporal convolutional and transformer-based models for closed-set ADL classification using RGB video only. mmaction2 employs a C2D-ResNet-50 backbone with an I3D-style classification head, where 2D convolutional filters are temporally extended to capture motion across short RGB video clips. Spatio-temporal features are aggregated via global average pooling across spatial and temporal dimensions, and the resulting feature representation is passed through a fully connected layer to predict a single ADL class from a predefined set of known activities, consistent with closed-set recognition protocols.

**Spatial temporal graph convolutional network (ST-GCN, unimodal)**: ST-GCN represents 3D skeleton joint data as nodes in a spatial-temporal graph where edges encode both anatomical joint connections and temporal continuity across frames. The network applies stacked spatial-temporal graph convolution layers to propagate information across joints and time, learning discriminative motion patterns from the skeleton sequence. For classification, the resulting graph features are aggregated using global average pooling across all joints and timesteps, and the pooled representation is passed through a fully connected layer to predict a single ADL class from a predefined set of known activities.

**The multimodal split attention fusion (MSAF) method (dual-modal)** [54]: MSAF combines features from parallel video and pose backbones using split-attention fusion blocks that enable cross-modal feature reweighting. RGB video features are extracted using a 3D convolutional video backbone, while skeleton-based motion features are extracted using a hierarchical co-occurrence network. MSAF blocks iteratively exchange information between the two modalities by learning attention weights that emphasize complementary features across streams. For classification, the fused feature representation is aggregated and passed through fully connected layers to predict one ADL class from a fixed set of known activities.

**POVNet (multimodal)** [24]: POVNet is our previously developed multimodal ADL classification architecture designed for closed-set recognition of a predefined set of known activities in socially assistive robotics. The method extracts scene and motion features from RGB video, user motion features from 3D skeleton trajectories, and semantic object cues via object detection. These modality-specific features are spatially aligned and integrated using a spatial mid-fusion method to form a fixed 128-dimensional ADL embedding, which is directly mapped to ADL class labels through supervised learning.

*Within-Domain Experiments with New Users*

A cross-subject training procedure was implemented using the ETRI-Activity-3D dataset, which included 3,712 samples of the 11 ADL task classes. We used the cross-subject (CS) split provided with ETRI, where approximately 2/3 of the subjects are used for training and 1/3 are used for testing. Within the training split, we further partitioned the data into training and validation sets using a 70/30 split.

The performance results are presented in Table 2. We also compared the model size and per-sample inference runtime of each approach. Model size reflects memory and deployment constraints relevant to SARs, while runtime represents computational efficiency during inference, which is important for real-time HRI. The same trained models and identical hardware/software conditions were used in the within-domain and out-of-domain comparisons.

Our POVNet+ had higher Precision (*P*), Recall (*R*), and *F1 score* compared to both the unimodal and dual-modal classification methods, as well as our previous multimodal POVNet method, while maintaining a compact model size (28.16 MB) and moderate inference runtime (0.623 s). We conducted nonparametric Friedman tests on the class predictions and determined that a statistically significant difference exists between class predictions for all methods: $Q = 318.851, p < 0.001$. Posthoc McNemar tests showed that our POVNet+ method had a statistically significant higher number of correct task predictions compared to: 1) SlowFast ($z = 15.727, p < 0.001$); 2) mmaction2 ($z =$

56.96, p = 0.001$); 3) ST-GCN ($z = 209.587, p < 0.001$); 4) MSAF ($z = 7.195, p = 0.0073$); and 5) POVNet ($z = 6.713, p = 0.0096$).

SlowFast uses only the video modality, while our POVNet+ model also includes 3D user pose data and ADL object location modalities. Although SlowFast has a substantially larger model size (128.67 MB) and slower inference runtime (1.019 s), the inclusion of the pose and object cues in POVNet+ enables improved classification accuracy for: 1) ADLs with distinctive user postures such as face washing), and 2) ADLs involving specific object interactions such as putting on shoes. mmaction2 is also a unimodal video baseline and therefore depends solely on RGB spatiotemporal features without explicit pose or object-location cues. While mmaction2 achieves competitive Precision, it has a relatively large model size (93.06 MB) and the slowest inference runtime (1.72 s), and is more sensitive to scene and motion bias when distinguishing ADLs with similar global motion but different object interactions (e.g., taking medicine versus drinking water). ST-GCN relies exclusively on pose features from the ETRI dataset, resulting in the smallest model size (10.07 MB) and fastest runtime (0.048 s); however, its lack of scene and object context leads to reduced classification performance for ADLs with similar motion patterns but differing object usage (e.g., drinking water versus eating food).

Compared to MSAF, POVNet+ includes object location features, resulting in higher classification accuracy for ADLs performed with specific objects (e.g., putting on shoes). Furthermore, the use of spatial mid-fusion allows POVNet+ to be robust to spatial variance, whereas the learned fusion method of MSAF does not explicitly contain a spatial reference. The slower runtime of POVNet+ compared to MSAF (0.623 s vs. 0.390 s) is due to the spatial mid-fusion requiring additional computational steps to manipulate the outputs of the backbone networks (e.g., reshaping and multiplying by absolute spatial position matrices), whereas the learned fusion approach of MSAF operates directly on backbone feature outputs. This added fusion stage also increases the model size of POVNet+ (28.16 mB) relative to POVNet (17.31 mB). Despite this overhead, POVNet+ achieves a higher overall *F1 score* while remaining suitable for multimodal SAR deployment. POVNet uses the same modalities as POVNet+, however, it employs a lower-resolution RGB backbone and a temporally independent pose backbone. This reduces its ability to accurately classify ADLs involving small-scale manipulations (e.g., taking medicine) or long temporal dependencies (e.g., folding laundry), despite its smaller model size (17.31 mB and comparable runtime (0.539 s).

TABLE 2. COMPARISON OF SOTA ADL CLASSIFICATION METHODS ON ETRI DATASET

| Method | Size (mB) | Run Time (s) | P | R | F1 Score |
|---|---|---|---|---|---|
| SlowFast | 128.67 | 1.019 | 0.843 | 0.836 | 0.836 |
| mmaction2 | 93.06 | 1.72 | 0.854 | 0.831 | 0.823 |
| ST-GCN | 10.07 | 0.048 | 0.755 | 0.741 | 0.742 |
| MSAF | 50.44 | 0.390 | 0.859 | 0.846 | 0.845 |
| POVNet | 17.31 | 0.539 | 0.859 | 0.850 | 0.849 |
| **POVNet+** | **28.16** | **0.623** | **0.877** | **0.866** | **0.865** |

*Out-of-Domain ADL Experiments with New Users*

For SARs to be deployed in real world environments, they need to be able to handle new environmental and user conditions. We created a new dataset using a different home-like environment and new users. This out-of-domain dataset was used to investigate the robustness of the above ADL classification methods. The dataset consists of 627 samples from four new users completing the same 11 ADL task classes in the ETRI dataset along with 2 new *unseen* ADL (NUADLs) task classes in a home-like living environment, Fig. 3. New classes include setting the table within the existing eating/drinking ADL class category and performing upper body arm exercises, a new exercise ADL class category.

Setting the table was selected to test robustness to combinations of known features while performing upper body arm exercises was selected to test robustness to new features.

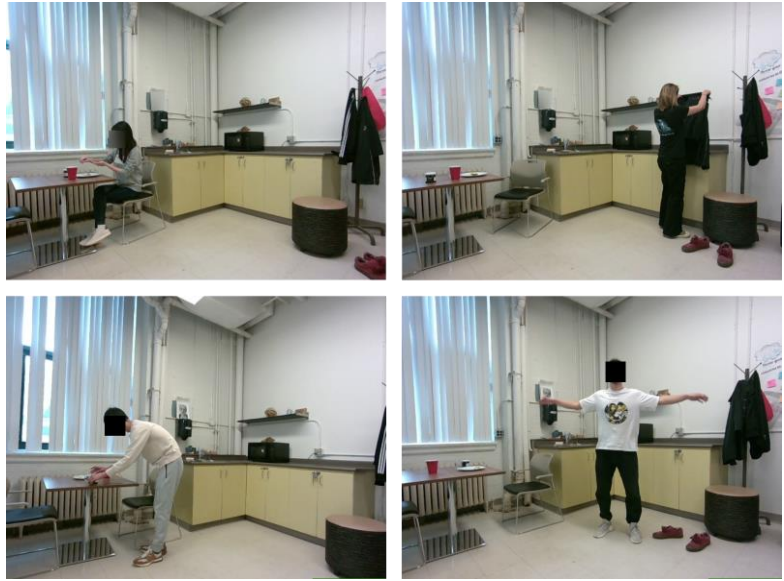

Figure 3. Examples from our Out-of-Domain ADL Dataset: a) taking medication, b) folding laundry, c) setting the table, and d) performing upper body arm exercises.

This new ADL dataset includes three distinct regions containing: 1) counter with sink, 2) table with chairs, and 3) coat rack and stools. All three regions are visible in each RGB video frame sample, which makes it more challenging to classify the ADL category using scene or object information. Samples consisting of RGB video and 3D user pose for the dataset were collected at 30 FPS throughout the duration of the ADL task. The ETRI dataset has temporally segmented samples of the user in the center of a single region while our out-of-domain samples include the beginning, middle, and end for each ADL task with the user in different frame locations for variance in spatial and temporal segmentation. Table 3 presents the comparison results. The accuracy for the NUADLs is defined as the total correct new ADL detections with respect to the total new ADL samples in the dataset.

TABLE 3. COMPARISON OF ADL CLASSIFICATION METHODS ON OUT-OF-DOMAIN ADL DATASET

| Method | Seen ADLs | | | NUADLs |
|---|---|---|---|---|
| | *P* | *R* | *F1 score* | Accuracy |
| SlowFast | 0.285 | 0.191 | 0.166 | 0.0 |
| mmaction2 | 0.152 | 0.206 | 0.107 | 0.0 |
| ST-GCN | 0.199 | 0.190 | 0.146 | 0.0 |
| MSAF | 0.077 | 0.113 | 0.064 | 0.0 |
| POVNet | 0.206 | 0.172 | 0.147 | 0.223 |
| **POVNet+** | **0.251** | **0.273** | **0.239** | **0.245** |

In general, our POVNet+ had higher *P*, *R*, and *F1 score* across the 11 ADLs, with the exception of SlowFast. Furthermore, it had a higher accuracy for the two *unseen* ADLs. A nonparametric Friedman test on the ADL task class predictions on the out-of-domain dataset for all models found a statistically significant difference between the models: $Q = 467.525$, $p < 0.001$. Posthoc McNemar tests determined that POVNet+ had a statistically significant higher number of correct task predictions than: 1) SlowFast ($z = 15.923$, $p < 0.001$), 2) mmaction2 ($z = 2.340$, $p < 0.001$), 3) ST-GCN ($z = 14.291$, $p < 0.001$), 4) MSAF ($z = 59.042$, $p < 0.001$), and 5) POVNet ($z = 28.474$, $p < 0.001$).

SlowFast exhibited higher Precision than POVNet+ for seen ADLs; however, this was primarily due to the strong class bias rather than robust generalization. Specifically, SlowFast predicted the taking medicine class for more than 50% of the out-of-domain samples, which inflated the Precision for this specific class, while significantly reducing Recall and *F1 score* across the remaining ADLs. This indicates that the unimodal video model over-predicted a single class in the new environment and relied heavily on appearance cues learned from the training videos, which did not transfer when the scene layout and object configurations changed. A similar degradation was observed for mmaction2, which also relies solely on RGB spatiotemporal features but uses fixed feature representations learned from the training distribution. In the unseen environment, variations in background layout, object placement, and task execution timing effected the learned motion–appearance correlations, reducing the model's ability to separate ADLs that share similar global motion but differ in task intent or object interaction. Consequently, mmaction2 exhibited reduced Recall and *F1 score* and was not able to detect the unseen ADL classes.

ST-GCN and MSAF had reduced performance in the out-of-domain setting. Namely, ST-GCN, while computationally efficient, does not incorporate scene or object context, limiting its ability to distinguish ADLs with similar motion patterns but different object interactions. MSAF incorporates video and pose modalities; however, its learned attention-based fusion does not explicitly encode spatial relationships between the user motion and objects, making it sensitive to changes in scene layout and user positioning. POVNet uses the same three modalities as POVNet+ but is a closed-set classifier without the additional robustness and embedding-based reasoning introduced in POVNet+; consequently, domain shift in scene layout and user positioning reduces its out-of-domain performance relative to POVNet+.

The improved NUADL accuracy of POVNet+ further demonstrates the advantage of its embedding-based formulation. By operating in a learned ADL embedding space rather than enforcing strict closed-set classification as done by the baseline classifiers SlowFast, ST-GCN, MSAF, POVNet, and mmaction2, POVNet+ can associate unseen activities with nearby known ADL clusters based on shared motion and object semantics. This capability is important for real-world SAR deployment, where robots must reason about previously unseen or atypically performed activities rather than solely relying on predefined action labels.

**HRI Experiments**

We conducted HRI experiments with a socially assistive robot to investigate the real-time performance of our entire DL architecture for autonomous ADL recognition and monitoring when providing multi-activity proactive assistive HRI. Namely, we considered different users performing multiple ADLs in sequence, encompassing distinct user behaviors during: 1) a *seen* ADL, 2) an *unseen* ADL, 3) an *atypically performed* ADL. Ten graduate students volunteered for the HRI experiment. All ten participants completed *seen* and *unseen* ADLs and five randomly selected participants from this group completed *atypically performed* ADLs. Participants were asked only to perform a specific ADL for each trial, with no further instructions. Ethics approval was obtained from the University of Toronto Office for Research Ethics, Research Ethics Board of Health Sciences, Protocol #44790.

*Set-Up and Procedure*

We used the same environment as in our out-of-domain experiments; however, we reconfigured the scene to increase clutter by moving the ADL objects closer to each other and adding background objects. The SAR Leia was positioned on top of the counter at approximately eye level to maintain line-of-sight of the user during ADL performance, Fig. 4. Leia's assistive behaviors included both speech and gestures, as shown in the video on our YouTube channel https://youtu.be/2VEQBh8b1-g.

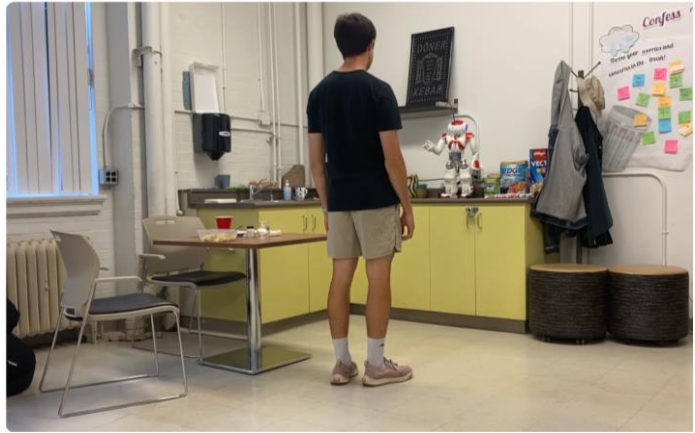

Figure 4. HRI Experiment Set-up with the Leia Robot.

*Seen and Unseen ADLs*

Four ADL tasks were randomly chosen from the following ADL categories: 1) a hygiene ADL task, located near the sink; 2) an eating/drinking ADL task, located at the table; 3) a dressing ADL task, located near the coat rack and stools; and 4) a new ADL. Leia then recognized the ADL being performed and provided the associated assistive instructions or identified that the ADL was new and provided a general reinforcement statement. Each trial consisted of one ADL being performed.

Results are presented in Table 4. Our POVNet+ method had an 80% success rate for both correctly recognizing *seen* ADL task classes and *unseen* ADLs as new activities. The lowest recognition rate for *seen* ADLs was for the eating/drinking ADL task classes (77.5%) which had high inter-class similarity, eating food with fork, drinking water, and taking medicine all had users move one of their hands towards their face while sitting down. Misclassification of *unseen* ADLs occurred when Leia identified them as a similar *seen* ADL due to the coexistence of an ADL feature, for example flossing was sometimes classified as brushing teeth due to the motion feature of a hand moving near the mouth, or an *unseen* ADL from a different ADL category due to a common feature between the

*unseen* ADL and the incorrect ADL category (e.g., setting the table associated with hygiene due to the user standing posture).

TABLE 4. HRI Results for Seen, Unseen, and Atypically Performed ADL

| ADL Type | ADL Category | ADL Task Class | ADLs Completed | # Successes | Success Rate |
|---|---|---|---|---|---|
| Seen | Hygiene | All Tasks | 40 | 32 | 80.0% |
| | Eating/Drinking | All Tasks | 40 | 31 | 77.5% |
| | Dressing | All Tasks | 40 | 33 | 82.5% |
| Unseen | Hygiene | Flossing Teeth | 10 | 9 | 90.0% |
| | Eating/Drinking | Setting the Table | 10 | 7 | 70.0% |
| | Dressing | Putting on Hat | 10 | 8 | 80.0% |
| Atypical | Hygiene | Washing Hands | 20 | 16 | 80.0% |
| | Eating/Drinking | Eating Food with a Fork | 20 | 17 | 85.0% |
| | Dressing | Putting on Jacket | 20 | 19 | 95.0% |

*ADLs Performed Atypically*

Participants first performed the three *seen* ADLs of washing hands, eating food with a fork, and putting on a jacket for five trials. These trials were used to determine the ADL embedding space for each specific user. Then, each participant was asked to alter their ADL behaviors in any manner they preferred for additional trials, defined here as *atypically performed ADL*. They were not given any specific instructions on how to modify their behaviors. *Atypically performed* ADL recognition and monitoring was defined to be successful when Leia was able to 1) correctly identify the ADL task class, 2) identify that the ADL type was *atypical*, and 3) inform the user (as shown in the video).

These results are also provided in Table IV. Our robot had an 86.7% detection success rate when an ADL was performed atypically. In general, *atypical* ADL behaviors observed during this experiment were: 1) applying soap only when handwashing and not performing rinsing/scrubbing, 2) holding the fork while at the table and being distracted by looking away or down at cell phone for a period, and 3) getting second arm stuck in the sleeve of the jacket.

*Discussions*

Our results validate the use of our POVNet+ method in successfully recognizing and classifying ADL types (seen, unseen and atypical) and the specific ADL task class, with an overall success rate of 81.9%. Classification accuracy can be improved by adding more classes in the object detection backbone such as napkins and medicine bottles to improve understanding of nuances between tasks that have similar user motion and background scene features. Future *unseen* classification may be extended to consider new ADL categories not included in the training data (e.g., meal preparation, exercise, etc.).

With respect to existing ADL recognition systems, classification has been reported only for *seen* ADLs with around 80% accuracy in offline testing with new users [55] and 60% in online testing [56] with new users. Our POVNet+ method obtains higher classification accuracy in real-time testing with a robot on a combination of new and known users.

**Conclusions**

In this paper, we present the development and integration of a deep learning architecture for SARs to recognize, monitor, and assist with multiple ADLs in real-time and in real scenarios. Our novel POVNet+ model introduces: 1) the use of both ADL and motion embedding spaces, and 2) the utilization of a novel user state estimation method with the motion embedding space to distinguish between a known ADL being performed (*seen*), a new (*unseen*) ADL, or a known ADL being performed atypically. Comparison experiments with our POVNet+ method validate its significantly higher classification accuracy compared to existing SOTA unimodal and dual-modal methods on the ETRI dataset and our own out-of-domain dataset. HRI experiments with our social robot Leia show the ability of our architecture to proactively initiate assistive HRI in a real environment by perceiving and responding to user ADL behaviors in real-time. This enables SARs to autonomously provide assistance without a priori knowledge of what ADL the user will perform. In general, our SAR ADL assistant architecture has the potential to be used for early detection of deteriorating physical or cognitive health in individuals by tracking changes in the ADL embedding space over time. Future work will include obtaining a larger dataset for training and testing in real-world environments with users who can benefit from a proactive SAR for ADL assistance.

## References

[1] Mlinac ME, Feng MC. Assessment of Activities of Daily Living, Self-Care, And Independence. Arch Clin Neuropsychol. 2016;31(6):506–516. https://doi.org/10.1093/arclin/acw049.
[2] Bloom DE, Canning D, Lubet A. Global Population Aging: Facts, Challenges, Solutions & Perspectives. Daedalus. 2015;144(2):80–92. https://doi.org/10.1162/DAED_a_00332.
[3] Edemekong PF, Bomgaars DL, Sukumaran S, Schoo C. Activities of Daily Living. In: StatPearls [Internet]. Treasure Island (FL): StatPearls Publishing; 2025 Jan–. 2025 May 4 [cited 2026 Jan 25]. Available from: https://pubmed.ncbi.nlm.nih.gov/29261878/.
[4] Vik K, Eide AH. The Exhausting Dilemmas Faced by Home-Care Service Providers When Enhancing Participation Among Older Adults Receiving Home Care. Scand J Caring Sci. 2012;26(3):528–536. https://doi.org/10.1111/j.1471-6712.2011.00960.x.
[5] Brownie S, Horstmanshof L, Garbutt R. Factors That Impact Residents' Transition and Psychological Adjustment to Long-Term Aged Care: A Systematic Literature Review. Int J Nurs Stud. 2014;51(12):1654–1666. https://doi.org/10.1016/j.ijnurstu.2014.04.011.
[6] Aspinal F, Glasby J, Rostgaard T, Tuntland H, Westendorp RG. New Horizons: Reablement—Supporting Older People Towards Independence. Age Ageing. 2016;45(5):572–576. https://doi.org/10.1093/ageing/afw094.
[7] Han DS, Chuang PW, Chiu EC. Effect of Home-Based Reablement Program on Improving Activities Of Daily Living For Patients with Stroke: A Pilot Study. Medicine (Baltimore). 2020;99(49):e23512. https://doi.org/10.1097/MD.0000000000023512.
[8] Pettersson C, Iwarsson S. Evidence-Based Interventions Involving Occupational Therapists Are Needed in Re-Ablement for Older Community-Living People: A Systematic Review. Br J Occup Ther. 2017;80(5):273–285. https://doi.org/10.1177/0308022617691537.
[9] Hjelle KM, Tuntland H, Førland O, Alvsvåg H. Driving Forces for Home-Based Reablement: A Qualitative Study of Older Adults' Experiences. Health Soc Care Community. 2017;25(5):1581–1589. https://doi.org/10.1111/hsc.12324.
[10] Vik K, Eide A. Older Adults Who Receive Home-Based Services, On the Verge of Passivity: The Perspective of Service Providers. Health Educ J. 2012;71(1):23–31. https://doi.org/10.1111/j.1748-3743.2011.00305.x.
[11] Robinson F, Cen Z, Naguib H, Nejat G. Socially Assistive Robotics and Wearable Sensors for Intelligent User Dressing Assistance. In: 2022 31st IEEE International Conference on Robot and Human Interactive Communication (RO-MAN); 2022 Aug; Napoli, Italy. IEEE; 2022. p. 829–836. https://doi.org/10.1109/RO-MAN53752.2022.9900778.
[12] Céspedes N, et al. A Socially Assistive Robot for Long-Term Cardiac Rehabilitation in the Real World. Front Neurorobot. 2021;15:633248. https://doi.org/10.3389/fnbot.2021.633248.

[13] McColl D, Louie WYG, Nejat G. Brian 2.1: A Socially Assistive Robot for the Elderly and Cognitively Impaired. IEEE Robot Autom Mag. 2013;20(1):74–83 https://doi.org/10.1109/MRA.2012.2229939.

[14] Shi C, Shiomi M, Kanda T, Ishiguro H, Hagita N. Measuring Communication Participation To Initiate Conversation in Human–Robot Interaction. Int J Soc Robot. 2015;7(5):889–910. https://doi.org/10.1007/s12369-015-0285-z.

[15] Clabaugh C, Matarić M. Escaping Oz: Autonomy in Socially Assistive Robotics. Annu Rev Control Robot Auton Syst. 2019;2:33–61. https://doi.org/10.1146/annurev-control-060117-104911.

[16] Bevilacqua R, Felici E, Cavallo F, Amabili G, Maranesi E. Designing Acceptable Robots for Assisting Older Adults: A Pilot Study on the Willingness to Interact. Int J Environ Res Public Health. 2021;18(20):10686. https://doi.org/10.3390/ijerph182010686.

[17] Tonmoy MTH, Mahmud S, Rahman AKMM, Amin MA, Ali AA. Hierarchical Self-Attention Based Autoencoder for Open-Set Human Activity Recognition. arXiv [Preprint]. 2021 Mar 7 [cited 2026 Jan 25]. Available from: https://arxiv.org/abs/2103.04279.

[18] Nan M, et al. Human Action Recognition for Social Robots. In: 2019 22nd International Conference on Control Systems and Computer Science (CSCS); 2019 May; Bucharest, Romania. IEEE; 2019. p. 675–681. https://doi.org/10.1109/CSCS.2019.00121.

[19] Cameron D, et al. The Effect of Social-Cognitive Recovery Strategies on Likability, Capability and Trust in Social Robots. Comput Hum Behav. 2021;114:106561. https://doi.org/10.1016/j.chb.2020.106561.

[20] Sasidharan S, Prabha P, Pasupuleti D, Das AM, Kapoor C, Manikutty G, Pankajakshan P, Rao B. Handwashing action detection system for an autonomous social robot. In: Proceedings of the IEEE Region 10 Conference (TENCON); Nov 2022; Hong Kong, China. IEEE; 2022. p. 1–6. https://doi.org/10.1109/TENCON55691.2022.9977684.

[21] Fasola J, Mataric MJ. Using Socially Assistive Human–Robot Interaction to Motivate Physical Exercise for Older Adults. Proc IEEE. 2012;100(8):2512–2526. https://doi.org/10.1109/JPROC.2012.2200539.

[22] Alves FR, Shao M, Nejat G. A Socially Assistive Robot to Facilitate and Assess Exercise Goals. In: Proceedings of the IEEE International Conference on Robotics and Automation Workshop on Mobile Robot Assistants for the Elderly; 2019; Montreal, QC, Canada. IEEE; 2019. p. 5. Available from: http://asblab.mie.utoronto.ca/sites/default/files/morobae_p10_alves.pdf.

[23] Astorga M, Cruz-Sandoval D, Favela J. A Social Robot to Assist in Addressing Disruptive Eating Behaviors by People with Dementia. Robotics. 2023;12(1):29. https://doi.org/10.3390/robotics12010029.

[24] Robinson F, Nejat G. A Deep Learning Human Activity Recognition Framework For Socially Assistive Robots to Support Reablement of Older Adults. In: 2023 IEEE International Conference on Robotics and Automation (ICRA); 2023 May; London, United Kingdom. IEEE; 2023. https://doi.org/10.1109/ICRA48891.2023.10161404.

[25] Montaño-Serrano VM, Jacinto-Villegas JM, Vilchis-González AH, Portillo-Rodríguez O. Artificial Vision Algorithms for Socially Assistive Robot Applications: A Review of the Literature. Sensors. 2021;21(17):5728. https://doi.org/10.3390/s21175728.

[26] Adama DA, Lotfi A, Langensiepen C, Lee K. Human Activities Transfer Learning for Assistive Robotics. In: Chao F, Schockaert S, Zhang Q, editors. Advances in Computational Intelligence Systems. Adv Intell Syst Comput. Cham: Springer International Publishing; 2018. p. 253–264. https://doi.org/10.1007/978-3-319-66939-7_22.

[27] Wu H, Pan W, Xiong X, Xu S. Human Activity Recognition Based on the Combined SVM and HMM. In: 2014 IEEE International Conference on Information and Automation (ICIA); 2014 Jul; Hailar, Hulun Buir, China. IEEE; 2014. p. 219–224. https://doi.org/10.1109/ICInfA.2014.6932656.

[28] Pashangpour S, Nejat G. The Future of Intelligent Healthcare: A Systematic Analysis and Discussion on the Integration and Impact of Robots using Large Language Models for Healthcare. Robotics. 2024;13(8):112. https://doi.org/10.3390/robotics13080112.

[29] Lisondra M, Benhabib B, Nejat G. Embodied AI With Foundation Models for Mobile Service Robots: A Systematic Review. arXiv [Preprint]. 2025 May 26 [cited 2026 Jan 25]. Available from: https://doi.org/10.48550/arXiv.2505.20503.

[30] Shu X, Yang J, Yan R, Song Y. Expansion-squeeze-excitation fusion network for elderly activity recognition. IEEE Trans Circuits Syst Video Technol. 2022;32(8):5281–5292. https://doi.org/10.1109/TCSVT.2022.3142771.

[31] Baselizadeh A, Uddin MZ, Khaksar W, Lindblom DS, Torresen J. PriMA-Care: Privacy-Preserving Multimodal Dataset for Human Activity Recognition in Care Robots. In: Companion of the 2024 ACM/IEEE International Conference on Human-Robot Interaction (HRI '24); 2024 Mar; New York,

NY, USA. Association for Computing Machinery; 2024. p. 233–237. https://doi.org/10.1145/3610978.3640701.

[32] Matarić MJ, Scassellati B. Socially Assistive Robotics. In: Siciliano B, Khatib O, editors. Springer handbook of robotics. Cham: Springer; 2016. https://doi.org/10.1007/978-3-319-32552-1_73.

[33] Poppe R. A Survey on Vision-Based Human Action Recognition. Image Vis Comput. 2010;28(6):976–990. https://doi.org/10.1016/j.imavis.2009.11.014.

[34] Santos L, Geminiani A, Schydlo P, Olivieri I, Santos-Victor J, Pedrocchi A. Design of a Robotic Coach for Motor, Social and Cognitive Skills Training Toward Applications With ASD Children. IEEE Trans Neural Syst Rehabil Eng. 2021;29:1223–1232. https://doi.org/10.1109/TNSRE.2021.3091320.

[35] Chen X, Kamavuako EN. Vision-Based Methods for Food and Fluid Intake Monitoring: A Literature Review. Sensors. 2023;23(13):6137. https://doi.org/10.3390/s23136137.

[36] McColl D, Nejat G. Meal-Time with a Socially Assistive Robot and Older Adults at a Long-Term Care Facility. J Hum Robot Interact. 2013;2(1):152–171. https://doi.org/10.5898/JHRI.2.1.McColl.

[37] Ferreira PJS, Cardoso JMP, Mendes-Moreira J. kNN Prototyping Schemes for Embedded Human Activity Recognition with Online Learning. Computers. 2020;9(4):96. https://doi.org/10.3390/computers9040096.

[38] Feichtenhofer C, Pinz A, Wildes RP. Spatiotemporal Multiplier Networks for Video Action Recognition. In: 2017 IEEE Conference on Computer Vision and Pattern Recognition (CVPR); 2017 Jul; Honolulu, HI, USA. IEEE; 2017. p. 7445–7454. https://doi.org/10.1109/CVPR.2017.787.

[39] Augustinov G, et al. Transformer-based recognition of activities of daily living from wearable sensor data. In: Proceedings of the 7th International Workshop on Sensor-Based Activity Recognition and Artificial Intelligence (iWOAR '22); 2023 Jan; New York, NY, USA. Association for Computing Machinery; 2023. p. 1–8. https://doi.org/10.1145/3558884.3558895.

[40] Fontana JM, Farooq M, Sazonov E. Automatic Ingestion Monitor: A Novel Wearable Device for Monitoring of Ingestive Behavior. IEEE Trans Biomed Eng. 2014;61(6):1772–1779. https://doi.org/10.1109/TBME.2014.2306773.

[41] Zhang Y, Maekawa T. InterHandNet: Capturing Two-Hand Interaction for Robust Hand-Washing Activity Recognition. In: 2025 IEEE International Conference on Pervasive Computing and Communications (PerCom); 2025 Mar; IEEE. p. 13–24. https://doi.org/10.1109/PerCom64205.2025.00021.

[42] Narasimhan S, Lisondra M, Wang H, Nejat G. SplatSearch: Instance Image Goal Navigation for Mobile Robots Using 3D Gaussian Splatting and Diffusion Models. arXiv [Preprint]. 2025 Nov 17 [cited 2026 Jan 25]. Available from: https://doi.org/10.48550/arXiv.2511.12972.

[43] De Lazzari D, et al. Real-Time Human Progress Estimation with Online Dynamic Time Warping For Collaborative Robotics. Front Robot AI. 2025;12:1623884. https://doi.org/10.3389/frobt.2025.1623884.

[44] De Lazzari D, et al. PACE: Proactive Assistance in Human–Robot Collaboration Through Action-Completion Estimation. In: 2025 IEEE International Conference on Robotics and Automation (ICRA); 2025 May; IEEE. p. 6725–6731. https://doi.org/10.1109/ICRA55743.2025.11127399.

[45] Rodomagoulakis I, et al. Multimodal Human Action Recognition in Assistive Human–Robot Interaction. In: 2016 IEEE International Conference on Acoustics, Speech and Signal Processing (ICASSP); 2016 Mar; IEEE. p. 2702–2706. https://doi.org/10.1109/ICASSP.2016.7472168.

[46] Nuitrack full body skeletal tracking software [Internet]. [cited 2026 Jan 25]. Available from: https://nuitrack.com/.

[47] Feichtenhofer C. X3D: Expanding Architectures for Efficient Video Recognition. arXiv [Preprint]. 2020 Apr 9 [cited 2026 Jan 25]. Available from: http://arxiv.org/abs/2004.04730.

[48] Yan S, Xiong Y, Lin D. Spatial Temporal Graph Convolutional Networks for Skeleton-Based Action Recognition. arXiv [Preprint]. 2018 Jan 25 [cited 2026 Jan 25]. Available from: http://arxiv.org/abs/1801.07455.

[49] Lei M, et al. YOLOv13: Real-Time Object Detection with Hypergraph-Enhanced Adaptive Visual Perception. arXiv [Preprint]. 2025 Sep 5 [cited 2026 Jan 25]. Available from: https://doi.org/10.48550/arXiv.2506.17733.

[50] Lin T-Y, et al. Microsoft COCO: Common Objects in Context. In: Fleet D, Pajdla T, Schiele B, Tuytelaars T, editors. Computer vision – ECCV 2014. Lect Notes Comput Sci. Cham: Springer International Publishing; 2014. p. 740–755. https://doi.org/10.1007/978-3-319-10602-1_48.

[51] Jang J, Kim D, Park C, Jang M, Lee J, Kim J. ETRI-Activity3D: a large-scale RGB-D dataset for robots to recognize daily activities of the elderly. In: 2020 IEEE/RSJ International Conference on Intelligent Robots and Systems (IROS); 2020 Oct; Las Vegas, NV, USA. IEEE; 2020. p. 10990–10997. https://doi.org/10.1109/IROS45743.2020.9341160.

[52] Feichtenhofer C, Fan H, Malik J, He K. SlowFast networks for video recognition. arXiv [Preprint]. 2019 Oct 29 [cited 2026 Jan 25]. Available from: http://arxiv.org/abs/1812.03982.

[53] MMAction Contributors. OpenMMLab's next generation video understanding toolbox and benchmark [Internet]. 2020 [cited 2026 Jan 25]. Available from: https://github.com/open-mmlab/mmaction2.

[54] Su L, Hu C, Li G, Cao D. MSAF: multimodal split attention fusion. arXiv [Preprint]. 2021 Jun 26 [cited 2026 Jan 25]. Available from: http://arxiv.org/abs/2012.07175.

[55] Massardi J, Gravel M, Beaudry E. PARC: A Plan and Activity Recognition Component for Assistive Robots. In: 2020 IEEE International Conference on Robotics and Automation (ICRA); 2020 May; Paris, France. IEEE; 2020. p. 3025–3031. https://doi.org/10.1109/ICRA40945.2020.9196856.

[56] Nasri N, et al. Assistive Robot with an AI-Based Application for the Reinforcement of Activities of Daily Living: Technical Validation with Users Affected by Neurodevelopmental Disorders. Appl Sci. 2022;12(19):9566. https://doi.org/10.3390/app12199566.